\documentclass[conference]{IEEEtran}
\IEEEoverridecommandlockouts
\usepackage{cite}
\usepackage{amsmath,amssymb,amsfonts}
\usepackage{algorithm}
\usepackage{algorithmic}
\usepackage{graphicx} 
\usepackage{booktabs} 
\usepackage{textcomp}
\usepackage{xcolor}
\usepackage{hyperref}
\usepackage{xcolor,tcolorbox}
\usepackage{amsmath}
\def\BibTeX{{\rm B\kern-.05em{\sc i\kern-.025em b}\kern-.08em
    T\kern-.1667em\lower.7ex\hbox{E}\kern-.125emX}}
\begin{document}

\title{Advancing Reasoning in Large Language Models: Promising Methods and Approaches}

\author{\IEEEauthorblockN{1\textsuperscript{st} Avinash Patil}
\IEEEauthorblockA{
\textit{Juniper Networks Inc.}\\
Sunnyvale, USA \\
patila@juniper.net \\ ORCID: 0009-0002-6004-370X}

\and
\IEEEauthorblockN{2\textsuperscript{nd} Aryan Jadon}
\IEEEauthorblockA{
\textit{Juniper Networks Inc.}\\
Sunnyvale, USA \\
aryanj@juniper.net \\
ORCID: 0000-0002-2991-9913}
}

\maketitle

\begin{abstract}
Large Language Models (LLMs) have succeeded remarkably in various natural language processing (NLP) tasks, yet their reasoning capabilities remain a fundamental challenge. While LLMs exhibit impressive fluency and factual recall, their ability to perform complex reasoning—spanning logical deduction, mathematical problem-solving, commonsense inference, and multi-step reasoning—often falls short of human expectations. This survey provides a comprehensive review of emerging techniques enhancing reasoning in LLMs. We categorize existing methods into key approaches, including prompting strategies (e.g., Chain-of-Thought reasoning, Self-Consistency, and Tree-of-Thought reasoning), architectural innovations (e.g., retrieval-augmented models, modular reasoning networks, and neuro-symbolic integration), and learning paradigms (e.g., fine-tuning with reasoning-specific datasets, reinforcement learning, and self-supervised reasoning objectives). Additionally, we explore evaluation frameworks used to assess reasoning in LLMs and highlight open challenges, such as hallucinations, robustness, and reasoning generalization across diverse tasks. By synthesizing recent advancements, this survey aims to provide insights into promising directions for future research and practical applications of reasoning-augmented LLMs.
\end{abstract}


\section{Introduction}
Large Language Models (LLMs) have revolutionized the field of Natural Language Processing (NLP), enabling breakthroughs in machine translation, text generation, question-answering, and other complex linguistic tasks. Despite their remarkable fluency and knowledge retention, these models often struggle with systematic reasoning—an essential capability for tasks requiring logical inference, problem-solving, and decision-making \cite{wu2024reasoning}. While LLMs can generate plausible-sounding responses, they frequently exhibit reasoning errors, inconsistencies, and hallucinations, limiting their reliability in critical domains such as scientific discovery, law, and medicine \cite{brown2020language} \cite{kojima2022large}.

Reasoning in AI broadly encompasses multiple cognitive processes, including deductive, inductive, abductive, and commonsense reasoning \cite{clark2021transformers, yang2024language, bhagavatula2019abductive, zhou2020evaluating, liu2016probabilistic}. Unlike retrieval-based knowledge synthesis, reasoning requires multi-step logical transformations, contextual generalization, and structured problem-solving. Classical AI approaches have addressed reasoning through rule-based symbolic systems  \cite{yager1984approximate} \cite{sun1995robust}, yet integrating such structured reasoning with the data-driven paradigm of LLMs remains an ongoing challenge. 

Recent research has explored diverse methodologies to enhance the reasoning abilities of LLMs. These approaches can categorized into three domains: (1) Prompting Strategies, such as Chain-of-Thought (CoT) reasoning \cite{wei2022chain}, Self-Consistency \cite{wang2022self}, and Tree-of-Thought \cite{yao2024tree} methods, which leverage structured prompts to guide step-by-step reasoning; (2) Architectural Innovations, including retrieval-augmented models \cite{lewis2020retrieval}, neuro-symbolic hybrid frameworks \cite{garcez2023neurosymbolic}, and modular reasoning architectures that integrate structured knowledge and logic \cite{NIPS2017_e6acf4b0}; and (3) Learning Paradigms, involving fine-tuning with specialized datasets \cite{zelikman2022star}, reinforcement learning for reasoning consistency \cite{guo2025deepseek}, and self-supervised objectives that encourage logical generalization \cite{talmor2020leap}. 

Among recent advancements, the newly released LLM DeepSeek-R1 \cite{guo2025deepseek} has demonstrated superior reasoning performance, particularly in complex domains such as mathematics and coding. By effectively simulating human-like analytical thinking, DeepSeek-R1 enhances multi-step reasoning in mathematical problem-solving, logical inference, and programming tasks, showcasing the potential of fine-tuned architectures and novel training paradigms to improve structured reasoning in LLMs.
This survey systematically reviews these advancements in LLM reasoning, assessing their effectiveness, limitations, and applications. It covers evaluation benchmarks, key challenges like adversarial robustness, cross-domain generalization, and reasoning biases. By synthesizing recent progress, we provide a comprehensive overview of promising techniques and future research directions.

The paper is structured as follows: Section 2 covers the foundations of reasoning, while Section 3 explores prompt-based reasoning enhancements. Section 4 discusses architectural innovations, and Section 5 examines learning-based approaches. Section 6 focuses on evaluation and benchmarking, Section 7 highlights challenges and open research directions, and Section 8 concludes the paper.

\section{Foundations of Reasoning in AI and LLMs}

\subsection{Definitions and Types of Reasoning}

Reasoning is the cognitive process of deriving conclusions from premises or evidence. It can classified into the following types:

\begin{itemize}
    \item \textbf{Deductive Reasoning}: Drawing specific conclusions from general premises. If the premises are true, the conclusion must be true. This method is fundamental in formal logic and automated theorem proving.
    \item \textbf{Inductive Reasoning}: Deriving general principles from specific examples or observations. This approach is common in machine learning for pattern recognition and forecasting.
    \item \textbf{Abductive Reasoning}: Inferring the most likely explanation for a given set of observations, frequently used in diagnostics and hypothesis formation.
    \item \textbf{Commonsense Reasoning}: Applying general world knowledge to infer reasonable conclusions is crucial for understanding implicit meanings in human communication.
    \item \textbf{Probabilistic Reasoning}: Handling uncertainty in logical inference using probability theory, often implemented in Bayesian networks and Markov models.
\end{itemize}

\subsection{Classical AI Approaches to Reasoning}

Traditional AI research has long focused on formal reasoning techniques incorporating structured knowledge representations. Some of the key classical approaches include \cite{yager1984approximate, sun1995robust}:

\begin{itemize}
    \item \textbf{Symbolic Logic}: Formal rule-based systems that use first-order logic (FOL) and propositional logic to derive conclusions.
    \item \textbf{Rule-Based Systems}: AI models that apply predefined rules to infer logical conclusions, used in expert systems and decision trees.
    \item \textbf{Knowledge Graphs}: Structured representations of entities and their relationships, supporting reasoning through graph traversal and inference mechanisms.
    \item \textbf{Automated Theorem Proving (ATP)}: Algorithms designed to prove mathematical theorems using logical deduction, such as the resolution principle in propositional logic.
    \item \textbf{Bayesian Networks}: Probabilistic graphical models that enable reasoning under uncertainty by representing dependencies between variables.
\end{itemize}

While these classical approaches provide strong logical foundations, they struggle with scalability and adaptability when applied to open-ended, unstructured problems such as natural language understanding.

\subsection{Reasoning in Large Language Models}

Large Language Models (LLMs) such as GPT-4, PaLM, and LLaMA utilize deep learning architectures, primarily transformers, to process and generate human-like text. However, their reasoning capabilities differ significantly from traditional AI approaches \cite{clark2021transformers, yang2024language, bhagavatula2019abductive, zhou2020evaluating, liu2016probabilistic}:

\begin{itemize}
    \item \textbf{Statistical Learning vs. Symbolic Logic}: Unlike symbolic AI, which follows explicit logical rules, LLMs learn probabilistic patterns in language data, making their reasoning implicit and non-deterministic.
    \item \textbf{Emergent Reasoning Abilities}: Studies suggest that scaling LLMs improves their ability to perform multi-step reasoning tasks despite the lack of explicit logical constraints.
    \item \textbf{Contextual and Prompt-Driven Reasoning}: LLMs rely heavily on context windows and external prompt engineering techniques (e.g., Chain-of-Thought prompting) to generate reasoned responses.
\end{itemize}

\subsection{Challenges of Reasoning in LLMs}

Despite their progress, LLMs face several challenges when it comes to robust and reliable reasoning \cite{huang2024survey, wang2024augmenting, lipton2018mythos}:

\begin{itemize}
    \item \textbf{Hallucinations}: LLMs sometimes generate plausible but incorrect information, leading to unreliable reasoning.
    \item \textbf{Lack of Explicit Memory}: Unlike knowledge graphs or rule-based systems, LLMs lack structured long-term memory, making reasoning consistency difficult.
    \item \textbf{Difficulty with Multi-Step Reasoning}: Although techniques like Chain-of-Thought prompting help, LLMs often fail to follow multi-step logical structures correctly.
    \item \textbf{Bias and Interpretability Issues}: Since LLMs train on vast text corpora, they inherit biases from data, which can influence reasoning outputs in unpredictable ways.
    \item \textbf{Limitations in Logical Deduction}: While LLMs excel at recognizing language patterns, they struggle with formal logic, mathematical proofs, and systematically verifying conclusions.
    \item \textbf{Limited Generalization Across Domains}: LLMs trained on diverse datasets still struggle with transferring reasoning skills across vastly different domains (e.g., legal reasoning vs. scientific inference).
\end{itemize}

\subsection{Bridging the Gap Between AI Reasoning and LLMs}

To enhance reasoning in LLMs, recent research \cite{lewis2020retrieval, garcez2023neurosymbolic, kumar2024training, guo2025deepseek} has explored hybrid models that integrate traditional reasoning techniques with deep learning. Key directions include :

\begin{itemize}
    \item \textbf{Fine-Tuning with Structured Reasoning Data}: Training LLMs on specialized datasets that explicitly focus on logical inference and mathematical problem-solving.
    \item \textbf{Retrieval-Augmented Reasoning}: Enhancing LLMs with knowledge retrieval mechanisms, allowing them to ground their responses in external facts.
    \item \textbf{Neuro-Symbolic AI}: Combining neural networks with symbolic reasoning frameworks to leverage the strengths of both approaches.
    \item \textbf{Self-Supervised and Reinforcement Learning Techniques}: Encouraging models to refine their reasoning through iterative self-training and reward mechanisms.
\end{itemize}

These advancements aim to push LLMs toward more reliable, explainable, and human-like reasoning capabilities.

\section{Prompting-Based Reasoning Enhancement}

\begin{figure*}[ht]
    \centering
    \includegraphics[width=0.75\textwidth]{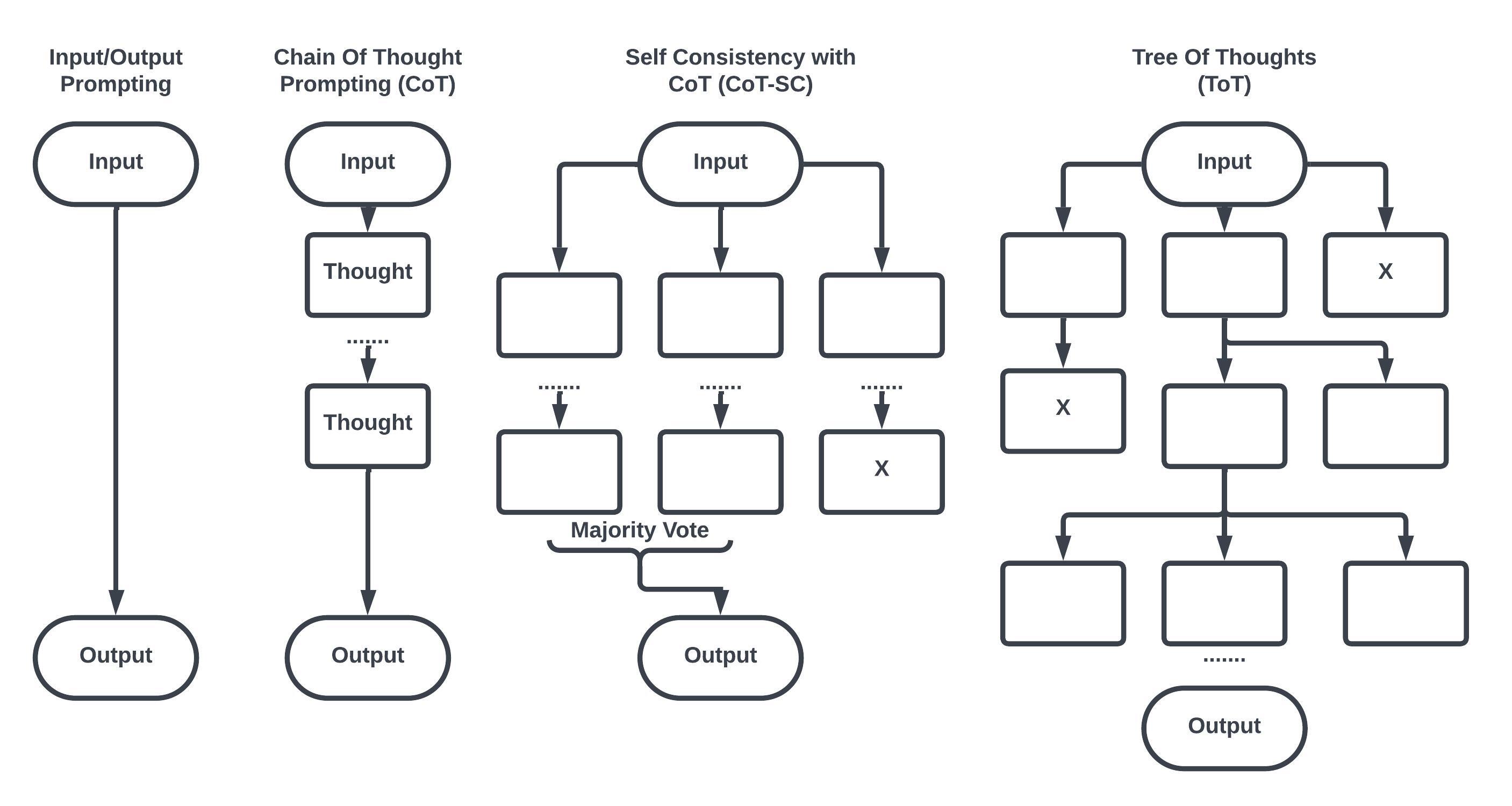} 
    \caption{Approaches to Prompting-Based Reasoning Enhancement.}
    \label{fig:tot}
\end{figure*}

Large Language Models (LLMs) demonstrate emergent reasoning through structured prompts, bypassing the need for fine-tuning \cite{brown2020language, wei2022emergent}. This section examines key prompting techniques, illustrated in Figure~\ref{fig:tot} and summarized in Table~\ref{tab:comparison}.

\subsection{Chain-of-Thought (CoT) Reasoning}

Chain-of-Thought (CoT) reasoning is a prompting technique used in large language models (LLMs) to improve their ability to solve complex reasoning problems. It involves breaking down a problem into a series of intermediate steps, allowing the model to reason more effectively and arrive at accurate conclusions \cite{wei2022chain}. This technique has been particularly effective for complex mathematical problem-solving, logical reasoning, and commonsense inference.

\begin{itemize}
    \item \textbf{Step-by-Step Reasoning}: Instead of answering immediately, the model generates a sequence of logical steps to work through the problem, improving accuracy in multi-step problem-solving.
    \item \textbf{Intermediate Reasoning}: The approach mimics human problem-solving by considering subproblems before reaching the final answer.
    \item \textbf{Performance Gains}: Studies show that CoT prompting improves performance on arithmetic and logical tasks compared to standard prompting \cite{wei2022chain}.
    \item \textbf{Limitations}: While CoT enhances interpretability, its effectiveness depends on prompt design and model size. In some cases, models may still generate incorrect intermediate steps \cite{wang2022self}.
\end{itemize}

\subsection{Self-Consistency Prompting}

Self-Consistency prompting is an advanced prompting technique that improves reasoning accuracy by generating multiple diverse reasoning paths and selecting the most consistent answer \cite{wang2022self}. This method is useful in complex reasoning tasks where a single Chain-of-Thought (CoT) might be prone to errors. This technique reduces variability in responses and increases accuracy by aggregating outputs.

\begin{itemize}
    \item \textbf{Multiple Reasoning Paths}: Instead of generating a single step-by-step solution, the model produces multiple different reasoning chains.
    \item \textbf{Diverse Thought Processes}: Each reasoning chain might follow a different logical approach, reducing biases in a single trajectory.
    \item \textbf{Majority Voting on Final Answer}: The final response is determined based on the most frequently occurring correct answer across generated samples.
\end{itemize}

\subsection{Tree-of-Thought (ToT) Reasoning}

Tree-of-Thought (ToT) reasoning is an advanced problem-solving framework that extends CoT reasoning by exploring multiple possible reasoning paths in a tree-like structure \cite{yao2024tree}. Instead of following a single linear reasoning path, ToT allows branching and evaluation at each step, leading to more robust and optimal solutions.

\begin{itemize}
    \item \textbf{Structured Exploration}: The model explores different paths in a tree-like structure, selecting the optimal reasoning route.
    \item \textbf{Decision Evaluation \& Pruning}: ToT reasoning is particularly effective in combinatorial and planning tasks.
    \item \textbf{Final Answer Selection}: The best reasoning path is selected based on a scoring or majority selection process \cite{yao2024tree}.
\end{itemize}

\subsection{Program-aided Language Models (PAL)}
Program-Aided Language Models (PAL) is a technique that enhances a language model’s reasoning capabilities by allowing it to call external computational tools—such as Python or symbolic solvers—to perform calculations, execute logic-based steps, or verify solutions. Instead of relying purely on internal token-based reasoning, PAL leverages external code execution for improved accuracy and reliability \cite{gao2023pal}.

\begin{itemize}
    \item \textbf{Execution-Based Verification}: The model generates reasoning steps in code format, which is executed to verify correctness.
    \item \textbf{Higher Accuracy in Mathematical Reasoning}: PAL has demonstrated superior performance in tasks requiring precise calculations.
    \item \textbf{Dependence on External Tools}: This approach requires integration with external computing environments, limiting its scalability \cite{gao2023pal}.
\end{itemize}

Empirical studies indicate that CoT and self-consistency prompting significantly improve reasoning performance, particularly in structured domains such as mathematics and logic \cite{wei2022chain, wang2022self}.

\begin{table*}[htbp]
    \caption{Comparison of Chain-of-Thought (CoT), Self-Consistency CoT (SC-CoT), Tree-of-Thought (ToT), and Program-Aided Language Models (PAL)}
    \label{tab:comparison}
    \centering
    \renewcommand{\arraystretch}{1.3}
    \begin{tabular}{|l|c|c|c|c|}
        \hline
        \textbf{Feature} & \textbf{CoT} & \textbf{SC-CoT} & \textbf{ToT} & \textbf{PAL} \\
        \hline
        \textbf{Reasoning Structure} & Linear step-by-step & Multiple CoTs with voting & Tree-like branching & Reasoning via code execution \\
        \hline
        \textbf{Error Handling} & Can propagate errors & Averages out mistakes & Prunes weak paths & Uses external execution \\
        \hline
        \textbf{Reasoning Diversity} & Single trajectory & Multiple independent paths & Branching & Uses symbolic computation or code \\
        \hline
        \textbf{Answer Selection} & Direct from one chain & Majority vote & Best branch selection & Extracted from program output \\
        \hline
        \textbf{Best Use Case} & Logical/math problems & High-confidence reasoning & Multi-step decision-making & Numerical/symbolic problems \\
        \hline
        \textbf{Execution Source} & Within LLM & Within LLM & Evaluates multiple paths & Uses external computation \\
        \hline
    \end{tabular}
\end{table*}

\section{Architectural Innovations for Enhanced Reasoning}

While prompting-based techniques have improved the reasoning capabilities of Large Language Models (LLMs), architectural innovations play a crucial role in enhancing their ability to perform structured and complex reasoning. This section explores various model architectures and modifications to improve logical inference, multi-step reasoning, and knowledge integration.

\subsection{Retrieval-Augmented Generation (RAG)}

Retrieval-Augmented Generation (RAG) is an AI framework that combines information retrieval with text generation. It enhances LLM reasoning by incorporating external knowledge sources. This approach improves the accuracy, relevance, and factual grounding of responses compared to relying solely on parametric memory\cite{lewis2020retrieval}.

\begin{itemize}
    \item \textbf{Query Processing}: The input query is processed and embedded into a vector space. The model searches for relevant documents using a retrieval system (e.g., dense passage retrieval, BM25). The retrieved documents are appended to the input. 
    \item \textbf{Knowledge-Enhanced Reasoning}: RAG-based models supplement their reasoning process based on both the query and retrieved information.
    \item \textbf{Reduction of Hallucinations}: By grounding responses in external data, RAG helps mitigate hallucinations often observed in purely generative models \cite{shuster2021retrieval}.
\end{itemize}

\subsection{Neuro-Symbolic Hybrid Models}

Neuro-Symbolic Hybrid Models combine neural networks (which excel at pattern recognition and learning from data) with symbolic AI (which enables reasoning, logic, and explicit knowledge representation). This fusion aims to create more explainable, generalizable, and robust AI systems \cite{garcez2023neurosymbolic}.

\begin{itemize}
    \item \textbf{Integration of Logic and Learning}: These models use neural networks to process unstructured text while employing symbolic logic for rule-based reasoning. Neural models extract features, while symbolic systems provide logical inference.
    \item \textbf{Enhanced Interpretability}: Symbolic components improve transparency, making reasoning steps more explainable. Rule-based systems, knowledge graphs, and formal logic enable structured reasoning.

\end{itemize}
\subsection{Memory-Augmented Neural Networks}

Memory-Augmented Neural Networks (MANNs) are AI models that integrate external memory with neural networks, enabling them to store, retrieve, and manipulate information dynamically. MANNs can read from and write to an external memory module, making them more adaptable for reasoning consistency over long sequences, lifelong learning, and few-shot learning tasks \cite{wang2024augmenting}.

\begin{itemize}
    \item \textbf{Controller (Neural Network Core)}: A neural network (typically an RNN or Transformer) that processes inputs and manages interactions with memory, determining when and how to read/write data.
    \item \textbf{External Memory Storage}: A structured memory component (e.g., a differentiable memory matrix or key-value store) that holds information over time. Unlike standard RNNs, which rely only on hidden states, MANNs explicitly retrieve and update memory.
    \item \textbf{Memory Access Mechanism}: Read/write operations in memory-augmented neural networks are typically differentiable, enabling gradient-based learning. Addressing mechanisms include content-based addressing, which retrieves memory by assessing similarity to stored data, and location-based addressing, which accesses memory based on positional or sequential order.
\end{itemize}

\subsection{Graph Neural Networks (GNNs) and Knowledge Graphs}
Graph Neural Networks (GNNs) offer a structured framework for reasoning by explicitly representing entities and their relationships, enabling logical inference and multi-hop question-answering.

\begin{itemize}
    \item \textbf{Structured Representation}: Graph Neural Networks are neural models designed to operate on graph-structured data. Unlike traditional deep learning models (which work on grids like images or sequences like text), GNNs can model complex relationships between interconnected entities \cite{ji2021survey}.
    \item \textbf{Reasoning over Knowledge Graphs}: Knowledge Graphs represent facts as entities and relationships in a structured format, typically as a triple (subject, predicate, object). When GNNs are applied to Knowledge Graphs, they enable reasoning, inference, and discovery of hidden relationships.\cite{hamilton2017inductive}.
    \item \textbf{Improvements in Explainability}: Knowledge graph-based reasoning enhances transparency by making inference paths explicit.
\end{itemize}

\subsection{Tool-Use and API Augmentations}

LLMs can be augmented with external tools and APIs to improve reasoning capabilities, leveraging specialized computational resources beyond language modeling \cite{schick2023toolformer}.

\begin{itemize}
    \item \textbf{Programmatic Reasoning}: Models invoke external calculators, theorem solvers, or search engines to validate reasoning steps.
    \item \textbf{Dynamic Data Integration}: As illustrated in Table~\ref{tab:api_types}, APIs enable real-time access to updated knowledge, improving the factual accuracy of reasoning \cite{mialon2023augmented}.
    \item \textbf{Limitations}: Dependence on external services introduces latency and requires access control mechanisms.
\end{itemize}

\begin{table}[htbp]
    \caption{Common API Types Used in AI Systems}
    \centering
    \renewcommand{\arraystretch}{1.3}
    \begin{tabular}{ll}
        \toprule
        \textbf{API Type} & \textbf{Example Use Cases} \\
        \midrule
        Web Search APIs & Bing, Google, Weather API for live information \\
        Computation APIs & Wolfram Alpha for advanced mathematical reasoning \\
        Database APIs & SQL, NoSQL for structured queries \\
        Cloud Services APIs & AWS, Google Cloud, OpenAI API for cloud \\
        Automation APIs & Zapier, IFTTT for automating workflows \\
        Financial APIs & Stock market APIs (Alpha Vantage, Yahoo Finance) \\
        \bottomrule
    \end{tabular}
    \label{tab:api_types}
\end{table}

Empirical results suggest that retrieval-augmented and neuro-symbolic models outperform standard transformer architectures in structured reasoning tasks \cite{lewis2020retrieval, garcez2023neurosymbolic}.

\section{Learning-Based Approaches for Reasoning}

Beyond prompting and architectural innovations, learning-based approaches are critical in improving reasoning capabilities in Large Language Models (LLMs). These approaches involve training paradigms such as fine-tuning with reasoning-specific datasets, reinforcement learning for consistency, and self-supervised learning for logical inference. This section explores various learning-based methodologies that enhance the reasoning abilities of LLMs.

\subsection{Supervised Fine-Tuning on Reasoning-Specific Datasets}

Fine-tuning LLMs on high-quality reasoning datasets allows models to improve their logical, mathematical, and commonsense reasoning capabilities.

\begin{itemize}
    \item \textbf{Mathematical and Logical Reasoning}: Fine-tuning on datasets such as MATH and GSM8K enhances mathematical problem-solving and logical inference skills \cite{cobbe2021gsm8k, hendrycks2measuring}.
    \item \textbf{Commonsense and Causal Reasoning}: Datasets like SWAG and Abductive NLI (aNLI) help models learn commonsense reasoning and abductive inference \cite{zellers2018swag, bhagavatula2019abductive}.
    \item \textbf{Scientific and Multi-Hop Reasoning}: Fine-tuning on datasets like ARC and HotpotQA improves multi-step reasoning and question-answering \cite{clark2018think, yang2018hotpotqa}.
\end{itemize}

While fine-tuning can significantly improve model performance, it requires careful dataset curation to prevent overfitting and ensure generalizability.

\subsection{Reinforcement Learning from Human Feedback}
Methods such as Reinforcement Learning from Human Feedback (RLHF) train models to align their reasoning with human preferences \cite{achiam2023gpt}. A PPO-based RLHF training algorithm is Algorithm~\ref{algo:ppo_rlhf}.

\begin{itemize}
    \item \textbf{Reward Models for Logical Consistency}: RLHF optimizes model outputs based on human evaluators' feedback, reducing errors in logical reasoning \cite{ouyang2022training}.
    \item \textbf{Reward Model (RM) Training}: Human annotators assess multiple model outputs based on preference. A dedicated neural network, known as the Reward Model, is trained on these rankings to capture human preferences. The models generate and assess their reasoning steps, refining correct solutions through iterative learning \cite{zelikman2022star}.
    \item \textbf{Reinforcement Learning via Proximal Policy Optimization (PPO)}: PPO, a reinforcement learning algorithm, is used to optimize the model while preventing drastic deviations from its base performance \cite{guo2025deepseek}.
\end{itemize}

\definecolor{lightblue}{rgb}{0.9, 0.95, 1} 
\begin{algorithm}
    \caption{RLHF Training Pipeline using PPO}
    \label{algo:ppo_rlhf}
    \begingroup
    \colorbox{lightblue}
    {
        \begin{minipage}{0.95\linewidth}
        \begin{algorithmic}[1]
        \STATE \textbf{Input:} Pre-trained language model $\mathcal{M}$, Supervised fine-tuning dataset $\mathcal{D}_{\text{SFT}}$, Reward model dataset $\mathcal{D}_{\text{RM}}$, Learning rate $\alpha$, Temperature $\tau$
        \STATE \textbf{Output:} RLHF-tuned model $\mathcal{M}_{\text{RLHF}}$
        \STATE
        
        \STATE \textbf{Step 1: Supervised Fine-Tuning (SFT)}
        \STATE Load pre-trained language model $\mathcal{M}$
        \STATE Load supervised fine-tuning dataset $\mathcal{D}_{\text{SFT}}$
        \STATE Train $\mathcal{M}$ on $\mathcal{D}_{\text{SFT}}$ using cross-entropy loss
        \STATE Save fine-tuned model as $\mathcal{M}_{\text{SFT}}$
        
        \STATE \textbf{Step 2: Train Reward Model}
        \STATE Initialize reward model $\mathcal{R}$
        \STATE Load ranked preference dataset $\mathcal{D}_{\text{RM}}$
        \STATE Train $\mathcal{R}$ to predict reward scores from human-ranked data
        \STATE Save trained reward model as $\mathcal{R}_{\text{trained}}$
        
        \STATE \textbf{Step 3: Reinforcement Learning with PPO}
        \STATE Initialize PPO agent using $\mathcal{M}_{\text{SFT}}$
        \STATE Set up PPO hyperparameters: batch size $B$, policy update steps $K$
        \FOR{each training iteration}
            \STATE Sample batch $\{x_i\} \in \mathcal{D}_{\text{SFT}}$
            \STATE Generate responses $y_i = \mathcal{M}_{\text{SFT}}(x_i)$
            \STATE Compute rewards $r_i = \mathcal{R}_{\text{trained}}(y_i)$
            \STATE Update policy $\pi_{\theta}$ using PPO objective:
            \[
            \mathcal{L}_{\text{PPO}} = \mathbb{E}_t \left[ \min \left( r_t(\theta) A_t, \text{clip}(r_t(\theta), 1 - \epsilon, 1 + \epsilon) A_t \right) \right]
            \]
            \STATE Perform gradient updates on $\mathcal{M}_{\text{SFT}}$
        \ENDFOR
        \STATE Save final RLHF-trained model as $\mathcal{M}_{\text{RLHF}}$
        \end{algorithmic}
        \end{minipage}
    }
    \endgroup
\end{algorithm}

\subsection{Self-Supervised and Contrastive Learning for Reasoning}

Self-supervised learning (SSL) and contrastive learning (CL) have gained traction as effective ways to train large-scale language models for reasoning tasks. Unlike supervised learning, which relies on human-labeled data, SSL and CL leverage inherent structures in data to create useful representations and improve reasoning capabilities \cite{talmor2020leap}.

\begin{itemize}
    \item \textbf{Contrastive Learning for Logical Inference}: By training models to distinguish between valid and invalid reasoning chains, contrastive learning improves logical consistency \cite{durkan2020contrastive}. Contrastive learning optimizes a contrastive loss, such as InfoNCE (Noise Contrastive Estimation) or Triplet Loss, which encourages correct reasoning pairs to have higher similarity scores. The InfoNCE loss function is defined as:

        \[L = - \sum_{i} \log \frac{\exp \left( \text{sim}(x_i, x_i^+) / \tau \right)}{\sum_{j} \exp \left(\text{sim}(x_i, x_j) / \tau \right)}\]
    
        where:
        \begin{itemize}
            \item \( x_i \) is the anchor sample,
            \item \( x_i^+ \) is the positive (similar) sample,
            \item \( x_j \) represents all samples in the denominator, including both positive and negative samples,
            \item \( \text{sim}(\cdot, \cdot) \) denotes a similarity function (e.g., cosine similarity),
            \item \( \tau \) is the temperature parameter.
        \end{itemize}
        
    \item \textbf{Self-Training with Synthetic Data}: Models generate synthetic reasoning paths and verify their correctness, iteratively refining their reasoning abilities \cite{zelikman2022star}.
    \item \textbf{Zero-Shot and Few-Shot Reasoning Improvement}: Self-supervised learning enhances a model's ability to generalize to novel reasoning tasks by enabling it to extract abstract reasoning patterns directly from raw data \cite{talmor2020leap}.
\end{itemize}

\subsection{Automated Verifiers and Critic Models}

To further enhance reasoning accuracy, LLMs can be paired with automated verifiers that critically assess their outputs \cite{tafjord2021proofwriter}.

\begin{itemize}
    \item \textbf{Secondary Verification Models}: A separate model evaluates the reasoning output of an LLM, filtering out incorrect inferences.
    \item \textbf{Formal Proof Checking}: Integration with theorem provers allows models to verify logical deductions rigorously \cite{first2023baldur}.
    \item \textbf{Limitations}: Automated verification remains challenging due to the difficulty of formalizing natural language reasoning.
\end{itemize}


\section{Evaluation and Benchmarking of Reasoning in LLMs}

Assessing the reasoning capabilities of Large Language Models (LLMs) requires systematic evaluation using standardized benchmarks and performance metrics. This section explores various evaluation methodologies, including reasoning benchmarks, key performance metrics, comparative analysis with human reasoning, and limitations of current evaluation strategies.

\subsection{Popular Reasoning Benchmarks}

Several benchmarks have been developed to assess different aspects of reasoning in LLMs, ranging from mathematical problem-solving to logical inference and commonsense reasoning.

\begin{itemize}
    \item \textbf{ARC (AI2 Reasoning Challenge)} – Measures commonsense and logical inference abilities by requiring multi-step reasoning across different knowledge domains \cite{clark2018think}.
    \item \textbf{LogiQA} – A dataset evaluating logical reasoning skills, particularly in deductive and abductive reasoning scenarios \cite{liu2021logiqa}.
    \item \textbf{GSM8K} – A dataset focused on grade-school mathematical reasoning problems, evaluating multi-step arithmetic reasoning capabilities \cite{cobbe2021gsm8k}.
    \item \textbf{MATH} – A benchmark designed to test models on high-school and competition-level mathematics, assessing formal mathematical reasoning \cite{hendrycks2measuring}.
    \item \textbf{BIG-Bench} – A broad dataset covering a variety of reasoning tasks, including logical reasoning, abstraction, and multi-hop inference \cite{srivastava2022bigbench}.
    \item \textbf{ProofWriter} – Evaluates the model’s ability to perform automated theorem proving and logical deduction \cite{tafjord2021proofwriter}.
    \item \textbf{HotpotQA} – A dataset focused on multi-hop question-answering requiring models to combine information from multiple sources for reasoning \cite{yang2018hotpotqa}.
    \item \textbf{HumanEval} – Evaluates the code-generating abilities of LLMs. It evaluates models' capacity to understand programming-related tasks and generate syntactically correct and functionally accurate code according to the provided specifications. \cite{chen2021evaluating}
    \item \textbf{ANLI (Adversarial NLI)} – Designed to test models on natural language inference through adversarially generated reasoning tasks \cite{nie2020adversarial}.
    \item \textbf{HellaSwag} – A benchmark designed to test commonsense natural language inference. It requires the model to predict the most likely ending of a sentence. \cite{zellers2018swag}.
    \item \textbf{Measuring Massive Multitask Language Understanding (MMLU)} – Evaluates general knowledge and problem-solving abilities across 57 subjects, including elementary mathematics, US history, computer science, and law. \cite{hendrycks2020measuring}.
\end{itemize}

\subsection{Metrics for Measuring Reasoning Performance}

Evaluating reasoning in LLMs involves multiple performance metrics tailored to different reasoning tasks.

\begin{itemize}
    \item \textbf{Accuracy}: Measures the correctness of model responses, often evaluated using \textit{Exact Match (EM)} and \textit{F1-score}, particularly in mathematical and logical reasoning tasks \cite{hendrycks2measuring}.
    \item \textbf{Logical Consistency}: Assesses whether a model's reasoning follows coherent logical steps across multiple queries. Often evaluated using theorem-proving datasets such as ProofWriter \cite{tafjord2021proofwriter}.
    \item \textbf{Explainability and Interpretability}: Evaluates the transparency of reasoning steps, especially in \textit{Chain-of-Thought (CoT)} models, by assessing the faithfulness of intermediate steps to the final answer \cite{wei2022chain}.
    \item \textbf{Self-Consistency}: Measures reasoning reliability by generating multiple independent responses to the same query and assessing agreement among outputs \cite{wang2022self}.
    \item \textbf{Multi-Hop Reasoning Score}: Used in datasets like HotpotQA to assess the model’s ability to integrate multiple pieces of evidence in complex reasoning tasks \cite{yang2018hotpotqa}.
    \item \textbf{Adversarial Robustness}: Tests the model's ability to maintain reasoning accuracy under adversarial perturbations, as evaluated in the ANLI dataset \cite{nie2020adversarial}.
    \item \textbf{Faithfulness and Verifiability}: Measures whether the model-generated reasoning steps can be independently verified and logically aligned with the final answer \cite{first2023baldur}.
    \item \textbf{Confidence Calibration}: Evaluates whether the model’s confidence in its predictions correlates with correctness, commonly measured using \textit{log-likelihood scores} and \textit{Brier Score} \cite{guo2017calibration}.
    \item \textbf{Reasoning Generalization}: Assesses how well the model performs on out-of-distribution (OOD) reasoning tasks, testing adaptability beyond its training data \cite{lake2018generalization}.
\end{itemize}

\section{Challenges and Open Research Directions}

Despite significant advancements in enhancing the reasoning capabilities of Large Language Models (LLMs), several challenges persist. These limitations hinder their reliability, robustness, and applicability in high-stakes domains. This section discusses key challenges and proposes open research directions to address them.

\subsection{Hallucinations and Misinformation}

One of the critical challenges in LLM reasoning is the generation of hallucinated or factually incorrect information \cite{huang2024survey}.

\begin{itemize}
    \item \textbf{Unverified Reasoning Steps}: LLMs sometimes generate plausible but incorrect reasoning chains, leading to logical inconsistencies \cite{mitchell2023debate}.
    \item \textbf{Fact-Checking Mechanisms}: Existing fact-checking techniques fail to filter misinformation in multi-step reasoning tasks \cite{mialon2023augmented}.
    \item \textbf{Open Research Direction}: Developing automated verifiers and integrating LLMs with structured databases to improve factual accuracy.
\end{itemize}

\subsection{Generalization Across Domains}

LLMs often struggle to generalize reasoning capabilities across different domains, limiting their adaptability to novel scenarios \cite{bommasani2021opportunities}.

\begin{itemize}
    \item \textbf{Domain-Specific Overfitting}: Fine-tuning on specific reasoning datasets may improve performance in targeted tasks but hinders adaptability to unseen domains \cite{hendrycks2measuring}.
    \item \textbf{Cross-Domain Transfer Learning}: Current transfer learning approaches have limitations in maintaining reasoning coherence across diverse contexts \cite{talmor2020leap}.
    \item \textbf{Open Research Direction}: Investigating meta-learning and continual learning strategies for cross-domain generalization.
\end{itemize}

\subsection{Robustness to Adversarial Attacks}

LLMs are vulnerable to adversarial perturbations that exploit reasoning weaknesses, leading to incorrect or misleading outputs \cite{nie2020adversarial}.

\begin{itemize}
    \item \textbf{Sensitivity to Input Variations}: Small modifications in prompts can lead to significantly different reasoning outputs, impacting reliability.
    \item \textbf{Adversarial Robustness Testing}: Existing benchmarks do not sufficiently evaluate LLMs against adversarial reasoning challenges \cite{ji2021survey}.
    \item \textbf{Open Research Direction}: Developing robust adversarial training techniques to improve resistance to input manipulations.
\end{itemize}

\subsection{Integrating Symbolic and Neural Reasoning}

LLMs rely on statistical pattern recognition rather than formal logical reasoning, leading to errors in complex inferencing tasks \cite{garcez2023neurosymbolic}.

\begin{itemize}
    \item \textbf{Limitations of Purely Neural Approaches}: LLMs struggle with structured logic, formal proofs, and abstract symbolic reasoning \cite{first2023baldur}.
    \item \textbf{Neuro-Symbolic AI}: Combining neural networks with symbolic reasoning frameworks enhances logical consistency and interpretability \cite{garcez2023neurosymbolic}.
    \item \textbf{Open Research Direction}: Advancing hybrid neuro-symbolic architectures for reasoning-augmented AI models.
\end{itemize}

\section{Conclusion}

Advancing reasoning in Large Language Models (LLMs) is a key milestone in AI development. Despite improvements in prompting, architecture, and learning-based methods, challenges remain in logical consistency, generalization, robustness, and interpretability. This survey reviews key approaches to enhancing LLM reasoning, categorized into prompting techniques, architectural innovations, and learning-driven strategies.

\subsection{Summary of Key Findings}

The key takeaways from this survey can be summarized as follows:

\begin{itemize}
    \item \textbf{Prompting Strategies}: Techniques such as Chain-of-Thought (CoT) prompting, Self-Consistency, and Tree-of-Thought (ToT) reasoning have shown significant improvements in structured problem-solving, logical inference, and multi-step reasoning \cite{wei2022chain, wang2022self, yao2024tree}.
    \item \textbf{Architectural Innovations}: Enhancements such as Retrieval-Augmented Generation (RAG), Neuro-Symbolic AI, Memory-Augmented Models, and Graph Neural Networks (GNNs) contribute to better structured and explainable reasoning \cite{lewis2020retrieval, garcez2023neurosymbolic}.
    \item \textbf{Learning-Based Approaches}: Fine-tuning on reasoning-specific datasets, Reinforcement Learning from Human Feedback (RLHF), self-supervised learning, and automated verifiers improve logical consistency and generalization \cite{hendrycks2measuring, ouyang2022training, zelikman2022star}.
    \item \textbf{Evaluation and Benchmarking}: Current benchmarks such as GSM8K, MATH, LogiQA, and ARC provide valuable insights into LLM reasoning capabilities, but existing evaluation methodologies require improvements in adversarial robustness and dynamic reasoning assessment \cite{cobbe2021gsm8k, hendrycks2measuring, liu2021logiqa}.
    \item \textbf{Challenges and Open Research Directions}: Key challenges include hallucinations, reasoning generalization, adversarial robustness, computational efficiency, ethical considerations, and the need for explainable reasoning models \cite{huang2024survey, bommasani2021opportunities, garcez2023neurosymbolic}.
\end{itemize}

\subsection{Final Thoughts}
The future of AI reasoning depends on developing models that generate fluent text while ensuring robust, verifiable, and adaptable reasoning across domains. Advancements in prompting, architecture, and learning can bring LLMs closer to human-like reasoning. However, addressing challenges requires collaboration among AI researchers, cognitive scientists, ethicists, and domain experts. The goal is to create AI systems that reason accurately, ethically, and transparently for safer real-world deployment.

\section{Acknowledgments}

We thank the research community for their contributions to reasoning in LLMs and developing benchmarking datasets. This survey has been informed by a wide range of studies, and we acknowledge the valuable work that has advanced the field.

\bibliographystyle{IEEEtran}
\nocite{*}
\bibliography{references}

\begin{thebibliography}{10}
\providecommand{\url}[1]{#1}
\csname url@samestyle\endcsname
\providecommand{\newblock}{\relax}
\providecommand{\bibinfo}[2]{#2}
\providecommand{\BIBentrySTDinterwordspacing}{\spaceskip=0pt\relax}
\providecommand{\BIBentryALTinterwordstretchfactor}{4}
\providecommand{\BIBentryALTinterwordspacing}{\spaceskip=\fontdimen2\font plus
\BIBentryALTinterwordstretchfactor\fontdimen3\font minus \fontdimen4\font\relax}
\providecommand{\BIBforeignlanguage}[2]{{%
\expandafter\ifx\csname l@#1\endcsname\relax
\typeout{** WARNING: IEEEtran.bst: No hyphenation pattern has been}%
\typeout{** loaded for the language `#1'. Using the pattern for}%
\typeout{** the default language instead.}%
\else
\language=\csname l@#1\endcsname
\fi
#2}}
\providecommand{\BIBdecl}{\relax}
\BIBdecl

\bibitem{wu2024reasoning}
Z.~Wu, L.~Qiu, A.~Ross, E.~Aky{\"u}rek, B.~Chen, B.~Wang, N.~Kim, J.~Andreas, and Y.~Kim, ``Reasoning or reciting? exploring the capabilities and limitations of language models through counterfactual tasks,'' in \emph{Proceedings of the 2024 Conference of the North American Chapter of the Association for Computational Linguistics: Human Language Technologies (Volume 1: Long Papers)}, 2024, pp. 1819--1862.

\bibitem{brown2020language}
T.~Brown \emph{et~al.}, ``Language models are few-shot learners,'' \emph{Advances in Neural Information Processing Systems}, 2020.

\bibitem{kojima2022large}
T.~Kojima, S.~S. Gu, M.~Reid, Y.~Matsuo, and Y.~Iwasawa, ``Large language models are zero-shot reasoners,'' \emph{Advances in neural information processing systems}, vol.~35, pp. 22\,199--22\,213, 2022.

\bibitem{clark2021transformers}
P.~Clark, O.~Tafjord, and K.~Richardson, ``Transformers as soft reasoners over language,'' in \emph{Proceedings of the Twenty-Ninth International Conference on International Joint Conferences on Artificial Intelligence}, 2021, pp. 3882--3890.

\bibitem{yang2024language}
Z.~Yang, L.~Dong, X.~Du, H.~Cheng, E.~Cambria, X.~Liu, J.~Gao, and F.~Wei, ``Language models as inductive reasoners,'' in \emph{Proceedings of the 18th Conference of the European Chapter of the Association for Computational Linguistics (Volume 1: Long Papers)}, 2024, pp. 209--225.

\bibitem{bhagavatula2019abductive}
C.~Bhagavatula, R.~L. Bras, C.~Malaviya, K.~Sakaguchi, A.~Holtzman, H.~Rashkin, D.~Downey, S.~W.-t. Yih, and Y.~Choi, ``Abductive commonsense reasoning,'' \emph{arXiv preprint arXiv:1908.05739}, 2019.

\bibitem{zhou2020evaluating}
X.~Zhou, Y.~Zhang, L.~Cui, and D.~Huang, ``Evaluating commonsense in pre-trained language models,'' in \emph{Proceedings of the AAAI Conference on Artificial Intelligence}, vol.~34, no.~05, 2020, pp. 9733--9740.

\bibitem{liu2016probabilistic}
Q.~Liu, H.~Jiang, A.~Evdokimov, Z.-H. Ling, X.~Zhu, S.~Wei, and Y.~Hu, ``Probabilistic reasoning via deep learning: Neural association models,'' \emph{arXiv preprint arXiv:1603.07704}, 2016.

\bibitem{yager1984approximate}
R.~R. Yager, ``Approximate reasoning as a basis for rule-based expert systems,'' \emph{IEEE Transactions on Systems, Man, and Cybernetics}, no.~4, pp. 636--643, 1984.

\bibitem{sun1995robust}
R.~Sun, ``Robust reasoning: integrating rule-based and similarity-based reasoning,'' \emph{Artificial Intelligence}, vol.~75, no.~2, pp. 241--295, 1995.

\bibitem{wei2022chain}
J.~Wei, X.~Wang, D.~Schuurmans, M.~Bosma, F.~Xia, E.~Chi, Q.~V. Le, D.~Zhou \emph{et~al.}, ``Chain-of-thought prompting elicits reasoning in large language models,'' \emph{Advances in neural information processing systems}, vol.~35, pp. 24\,824--24\,837, 2022.

\bibitem{wang2022self}
X.~Wang \emph{et~al.}, ``Self-consistency improves chain of thought reasoning in language models,'' \emph{arXiv preprint arXiv:2203.11171}, 2022.

\bibitem{yao2024tree}
S.~Yao, D.~Yu, J.~Zhao, I.~Shafran, T.~Griffiths, Y.~Cao, and K.~Narasimhan, ``Tree of thoughts: Deliberate problem solving with large language models,'' \emph{Advances in Neural Information Processing Systems}, vol.~36, 2024.

\bibitem{lewis2020retrieval}
P.~Lewis \emph{et~al.}, ``Retrieval-augmented generation for knowledge-intensive nlp tasks,'' \emph{Advances in Neural Information Processing Systems}, 2020.

\bibitem{garcez2023neurosymbolic}
A.~d. Garcez and L.~C. Lamb, ``Neurosymbolic ai: The 3 rd wave,'' \emph{Artificial Intelligence Review}, vol.~56, no.~11, pp. 12\,387--12\,406, 2023.

\bibitem{NIPS2017_e6acf4b0}
A.~Santoro, D.~Raposo, D.~G. Barrett, M.~Malinowski, R.~Pascanu, P.~Battaglia, and T.~Lillicrap, ``A simple neural network module for relational reasoning,'' in \emph{Advances in Neural Information Processing Systems}, I.~Guyon, U.~V. Luxburg, S.~Bengio, H.~Wallach, R.~Fergus, S.~Vishwanathan, and R.~Garnett, Eds., vol.~30.\hskip 1em plus 0.5em minus 0.4em\relax Curran Associates, Inc., 2017.

\bibitem{zelikman2022star}
E.~Zelikman, Y.~Wu, J.~Mu, and N.~Goodman, ``Star: Bootstrapping reasoning with reasoning,'' \emph{Advances in Neural Information Processing Systems}, vol.~35, pp. 15\,476--15\,488, 2022.

\bibitem{guo2025deepseek}
D.~Guo, D.~Yang, H.~Zhang, J.~Song, R.~Zhang, R.~Xu, Q.~Zhu, L.~Li, Z.~Shao, P.~Wang \emph{et~al.}, ``Deepseek-r1: Incentivizing reasoning capability in llms via reinforcement learning,'' \emph{arXiv preprint arXiv:2501.12948}, 2025.

\bibitem{talmor2020leap}
A.~Talmor, O.~Tafjord, P.~Clark, Y.~Goldberg, and J.~Berant, ``Leap-of-thought: Teaching pre-trained models to systematically reason over implicit knowledge,'' \emph{Advances in Neural Information Processing Systems}, vol.~33, pp. 20\,227--20\,237, 2020.

\bibitem{huang2024survey}
L.~Huang, W.~Yu, W.~Ma, W.~Zhong, Z.~Feng, H.~Wang, Q.~Chen, W.~Peng, X.~Feng, B.~Qin \emph{et~al.}, ``A survey on hallucination in large language models: Principles, taxonomy, challenges, and open questions,'' \emph{ACM Transactions on Information Systems}, 2024.

\bibitem{wang2024augmenting}
W.~Wang, L.~Dong, H.~Cheng, X.~Liu, X.~Yan, J.~Gao, and F.~Wei, ``Augmenting language models with long-term memory,'' \emph{Advances in Neural Information Processing Systems}, vol.~36, 2024.

\bibitem{lipton2018mythos}
Z.~C. Lipton, ``The mythos of model interpretability: In machine learning, the concept of interpretability is both important and slippery.'' \emph{Queue}, vol.~16, no.~3, pp. 31--57, 2018.

\bibitem{kumar2024training}
A.~Kumar, V.~Zhuang, R.~Agarwal, Y.~Su, J.~D. Co-Reyes, A.~Singh, K.~Baumli, S.~Iqbal, C.~Bishop, R.~Roelofs \emph{et~al.}, ``Training language models to self-correct via reinforcement learning,'' \emph{arXiv preprint arXiv:2409.12917}, 2024.

\bibitem{wei2022emergent}
J.~Wei \emph{et~al.}, ``Emergent abilities of large language models,'' \emph{arXiv preprint arXiv:2206.07682}, 2022.

\bibitem{gao2023pal}
L.~Gao, A.~Madaan, S.~Zhou, U.~Alon, P.~Liu, Y.~Yang, J.~Callan, and G.~Neubig, ``Pal: Program-aided language models,'' in \emph{International Conference on Machine Learning}.\hskip 1em plus 0.5em minus 0.4em\relax PMLR, 2023, pp. 10\,764--10\,799.

\bibitem{shuster2021retrieval}
K.~Shuster, S.~Poff, M.~Chen, D.~Kiela, and J.~Weston, ``Retrieval augmentation reduces hallucination in conversation,'' in \emph{Findings of the Association for Computational Linguistics: EMNLP 2021}, 2021, pp. 3784--3803.

\bibitem{ji2021survey}
S.~Ji, S.~Pan, E.~Cambria, P.~Marttinen, and S.~Y. Philip, ``A survey on knowledge graphs: Representation, acquisition, and applications,'' \emph{IEEE transactions on neural networks and learning systems}, vol.~33, no.~2, pp. 494--514, 2021.

\bibitem{hamilton2017inductive}
W.~L. Hamilton \emph{et~al.}, ``Inductive representation learning on large graphs,'' \emph{Advances in Neural Information Processing Systems}, 2017.

\bibitem{schick2023toolformer}
T.~Schick, J.~Dwivedi-Yu, R.~Dess{\`\i}, R.~Raileanu, M.~Lomeli, E.~Hambro, L.~Zettlemoyer, N.~Cancedda, and T.~Scialom, ``Toolformer: Language models can teach themselves to use tools,'' \emph{Advances in Neural Information Processing Systems}, vol.~36, pp. 68\,539--68\,551, 2023.

\bibitem{mialon2023augmented}
\BIBentryALTinterwordspacing
G.~Mialon, R.~Dessi, M.~Lomeli, C.~Nalmpantis, R.~Pasunuru, R.~Raileanu, B.~Roziere, T.~Schick, J.~Dwivedi-Yu, A.~Celikyilmaz, E.~Grave, Y.~LeCun, and T.~Scialom, ``Augmented language models: a survey,'' \emph{Transactions on Machine Learning Research}, 2023, survey Certification. [Online]. Available: \url{https://openreview.net/forum?id=jh7wH2AzKK}
\BIBentrySTDinterwordspacing

\bibitem{cobbe2021gsm8k}
K.~Cobbe \emph{et~al.}, ``Training verifiers to solve math word problems,'' \emph{arXiv preprint arXiv:2110.14168}, 2021.

\bibitem{hendrycks2measuring}
D.~Hendrycks, C.~Burns, S.~Kadavath, A.~Arora, S.~Basart, E.~Tang, D.~Song, and J.~Steinhardt, ``Measuring mathematical problem solving with the math dataset,'' \emph{Sort}, vol.~2, no.~4, pp. 0--6, 2021.

\bibitem{zellers2018swag}
R.~Zellers, Y.~Bisk, R.~Schwartz, and Y.~Choi, ``Swag: A large-scale adversarial dataset for grounded commonsense inference,'' \emph{arXiv preprint arXiv:1808.05326}, 2018.

\bibitem{clark2018think}
P.~Clark, I.~Cowhey, O.~Etzioni, T.~Khot, A.~Sabharwal, C.~Schoenick, and O.~Tafjord, ``Think you have solved question answering? try arc, the ai2 reasoning challenge,'' \emph{arXiv preprint arXiv:1803.05457}, 2018.

\bibitem{yang2018hotpotqa}
Z.~Yang, P.~Qi, S.~Zhang, Y.~Bengio, W.~Cohen, R.~Salakhutdinov, and C.~D. Manning, ``Hotpotqa: A dataset for diverse, explainable multi-hop question answering,'' in \emph{Proceedings of the 2018 Conference on Empirical Methods in Natural Language Processing}, 2018, pp. 2369--2380.

\bibitem{achiam2023gpt}
J.~Achiam, S.~Adler, S.~Agarwal, L.~Ahmad, I.~Akkaya, F.~L. Aleman, D.~Almeida, J.~Altenschmidt, S.~Altman, S.~Anadkat \emph{et~al.}, ``Gpt-4 technical report,'' \emph{arXiv preprint arXiv:2303.08774}, 2023.

\bibitem{ouyang2022training}
L.~Ouyang, J.~Wu, X.~Jiang, D.~Almeida, C.~Wainwright, P.~Mishkin, C.~Zhang, S.~Agarwal, K.~Slama, A.~Ray \emph{et~al.}, ``Training language models to follow instructions with human feedback,'' \emph{Advances in neural information processing systems}, vol.~35, pp. 27\,730--27\,744, 2022.

\bibitem{durkan2020contrastive}
C.~Durkan, I.~Murray, and G.~Papamakarios, ``On contrastive learning for likelihood-free inference,'' in \emph{International conference on machine learning}.\hskip 1em plus 0.5em minus 0.4em\relax PMLR, 2020, pp. 2771--2781.

\bibitem{tafjord2021proofwriter}
O.~Tafjord, B.~Dalvi, and P.~Clark, ``Proofwriter: Generating implications, proofs, and abductive statements over natural language,'' in \emph{Findings of the Association for Computational Linguistics: ACL-IJCNLP 2021}, 2021, pp. 3621--3634.

\bibitem{first2023baldur}
E.~First, M.~N. Rabe, T.~Ringer, and Y.~Brun, ``Baldur: Whole-proof generation and repair with large language models,'' in \emph{Proceedings of the 31st ACM Joint European Software Engineering Conference and Symposium on the Foundations of Software Engineering}, 2023, pp. 1229--1241.

\bibitem{liu2021logiqa}
J.~Liu, L.~Cui, H.~Liu, D.~Huang, Y.~Wang, and Y.~Zhang, ``Logiqa: a challenge dataset for machine reading comprehension with logical reasoning,'' in \emph{Proceedings of the Twenty-Ninth International Conference on International Joint Conferences on Artificial Intelligence}, 2021, pp. 3622--3628.

\bibitem{srivastava2022bigbench}
A.~Srivastava \emph{et~al.}, ``Beyond the imitation game: Quantifying and extrapolating the capabilities of language models,'' \emph{arXiv preprint arXiv:2206.04615}, 2022.

\bibitem{chen2021evaluating}
M.~Chen, J.~Tworek, H.~Jun, Q.~Yuan, H.~P. D.~O. Pinto, J.~Kaplan, H.~Edwards, Y.~Burda, N.~Joseph, G.~Brockman \emph{et~al.}, ``Evaluating large language models trained on code,'' \emph{arXiv preprint arXiv:2107.03374}, 2021.

\bibitem{nie2020adversarial}
Y.~Nie, A.~Williams, E.~Dinan, M.~Bansal, J.~Weston, and D.~Kiela, ``Adversarial nli: A new benchmark for natural language understanding,'' in \emph{Proceedings of the 58th Annual Meeting of the Association for Computational Linguistics}.\hskip 1em plus 0.5em minus 0.4em\relax Association for Computational Linguistics, 2020.

\bibitem{hendrycks2020measuring}
D.~Hendrycks, C.~Burns, S.~Basart, A.~Zou, M.~Mazeika, D.~Song, and J.~Steinhardt, ``Measuring massive multitask language understanding,'' \emph{arXiv preprint arXiv:2009.03300}, 2020.

\bibitem{guo2017calibration}
C.~Guo, G.~Pleiss, Y.~Sun, and K.~Q. Weinberger, ``On calibration of modern neural networks,'' in \emph{International Conference on Machine Learning (ICML)}, 2017, pp. 1321--1330.

\bibitem{lake2018generalization}
B.~Lake and M.~Baroni, ``Generalization without systematicity: On the compositional skills of sequence-to-sequence recurrent networks,'' in \emph{International conference on machine learning}.\hskip 1em plus 0.5em minus 0.4em\relax PMLR, 2018, pp. 2873--2882.

\bibitem{mitchell2023debate}
M.~Mitchell and D.~C. Krakauer, ``The debate over understanding in ai’s large language models,'' \emph{Proceedings of the National Academy of Sciences}, vol. 120, no.~13, p. e2215907120, 2023.

\bibitem{bommasani2021opportunities}
R.~Bommasani, D.~A. Hudson, E.~Adeli, R.~Altman, S.~Arora, S.~von Arx, M.~S. Bernstein, J.~Bohg, A.~Bosselut, E.~Brunskill \emph{et~al.}, ``On the opportunities and risks of foundation models,'' \emph{arXiv preprint arXiv:2108.07258}, 2021.

\end{thebibliography}
\vspace{12pt}

\end{document}